\documentclass[conference]{IEEEtran}
\IEEEoverridecommandlockouts
\usepackage{amsmath,amssymb,amsfonts}
\usepackage{graphicx}
\usepackage{textcomp}
\usepackage{xcolor}
\usepackage{bbm}

\usepackage{pgfplots} 
\pgfplotsset{compat=newest} 
\pgfplotsset{plot coordinates/math parser=false} 
\newlength\figureheight 
\newlength\figurewidth 

\usepackage{standalone}

\pgfplotsset{every axis legend/.append style={legend cell align=left}}

\usepackage{algorithm}% http://ctan.org/pkg/algorithm
\usepackage{algpseudocode}% http://ctan.org/pkg/algorithmicx

\usepackage[backend=bibtex, style=ieee, maxcitenames=2, mincitenames=1,doi=false]{biblatex}
\usepackage{todonotes}
\usepackage{xargs}
\newcommandx{\chris}[2][1=]{\todo[inline, backgroundcolor=yellow!25,bordercolor=red,#1]{#2 -CL}}
\newcommandx{\mykel}[2][1=]{\todo[inline, backgroundcolor=green!25,bordercolor=red,#1]{#2 -MK}}

\def\BibTeX{{\rm B\kern-.05em{\sc i\kern-.025em b}\kern-.08em
    T\kern-.1667em\lower.7ex\hbox{E}\kern-.125emX}}
    
\usepackage[keeplastbox]{flushend}
    
\begin{document}

\title{Runtime Safety Assurance \\ Using Reinforcement Learning\\
}

\author{\IEEEauthorblockN{Christopher Lazarus}
\IEEEauthorblockA{
\textit{Stanford University}\\
Stanford, CA \\
clazarus@stanford.edu}
\and
\IEEEauthorblockN{James G. Lopez}
\IEEEauthorblockA{\textit{GE Research} \\
Niskayuna, NY \\
lopezj@ge.com}
\and
\IEEEauthorblockN{Mykel J. Kochenderfer}
\IEEEauthorblockA{
\textit{Stanford University}\\
Stanford, CA \\
mykel@stanford.edu}
\iffalse
\and
\IEEEauthorblockN{3\textsuperscript{rd} Given Name Surname}
\IEEEauthorblockA{\textit{dept. name of organization (of Aff.)} \\
\textit{name of organization (of Aff.)}\\
City, Country \\
email address or ORCID}
\and
\IEEEauthorblockN{4\textsuperscript{th} Given Name Surname}
\IEEEauthorblockA{\textit{dept. name of organization (of Aff.)} \\
\textit{name of organization (of Aff.)}\\
City, Country \\
email address or ORCID}
\and
\IEEEauthorblockN{5\textsuperscript{th} Given Name Surname}
\IEEEauthorblockA{\textit{dept. name of organization (of Aff.)} \\
\textit{name of organization (of Aff.)}\\
City, Country \\
email address or ORCID}
\and
\IEEEauthorblockN{6\textsuperscript{th} Given Name Surname}
\IEEEauthorblockA{\textit{dept. name of organization (of Aff.)} \\
\textit{name of organization (of Aff.)}\\
City, Country \\
email address or ORCID}
\fi
}

\maketitle

\begin{abstract}
%Traditionally, controllers for aerospace and other cyber-physical systems (CPS) were developed by human engineers in a rigorous manner. 
%Safety requirements were specified and encoded as invariants that the systems were designed to obey at all times while also fulfilling their tasks efficiently and reliably. 
%However, verifying a controller for a broad array of scenarios can be extremely costly. 
%In the aeronautics domain, 
The airworthiness and safety of a non-pedigreed autopilot must be verified, but the cost to formally do so can be prohibitive.
We can bypass formal verification of non-pedigreed components by incorporating Runtime Safety Assurance (RTSA) as mechanism to ensure safety.
RTSA consists of a meta-controller that observes the inputs and outputs of a non-pedigreed component and verifies formally specified behavior as the system operates. When the system is triggered, a verified recovery controller is deployed. 
Recovery controllers are designed to be safe but very likely disruptive to the operational objective of the system, and thus RTSA systems must balance safety and efficiency. 
The objective of this paper is to design a meta-controller capable of identifying unsafe situations with high accuracy. High dimensional and non-linear dynamics in which modern controllers are deployed along with the black-box nature of the nominal controllers make this a difficult problem.
Current approaches rely heavily on domain expertise and human engineering. We frame the design of RTSA with the  Markov decision process (MDP) framework and use reinforcement learning (RL) to solve it.
Our learned meta-controller consistently exhibits superior performance in our experiments compared to our baseline, human engineered approach.
\end{abstract}


\section{Introduction}
\label{sec:rtsa}
%\label{sec:introduction}
%\textcolor{blue}{Explain why current trends lead to the adoption of controllers with non-pedigreed components: either with ML components or commercial off the shelf. This creates the need to guarantee safety with black box components. One way to do it during runtime using RTSA. RTSA has been proposed before and can induce safety even when black boxes are present. Here we present an approach to design the RTSA system: given a nominal controller (non-pedigreed) and a recovery system the system has to decide when to switch to  the recovery system. This can be a complicated task and a way to solve this optimally is using Reinforcement Learning. The final model has to be efficient and transparent so that restricts the type of policies that can be used. We postulate that determining when to interrupt a nominal controller is an easier task than controlling the system itself so a simple linear model should suffice while providing transparency and trustworthyness.}

There is a growing need to enable the deployment of controllers with black-box components in systems with complex dynamics. 
Unfortunately, the cost to formally verify a non-pedigreed or black-box autopilot for a variety of vehicle types and use cases is generally prohibitive.
An alternative is to bound the flight behavior of unmanned aircraft during operation to comply with safety constraints by mitigating the risk of uncontrolled flight beyond authorized conditions. 
Runtime Safety Assurance (RTSA) aims to do this as a runtime monitoring safeguard that is capable of switching to a safe recovery controller if the vehicle is at risk of operating unsafely.
This idea has been used in the context of software engineering \cite{setodynamic} and has also been proposed in the aerospace domain where a trusted interpretable simple controller was used to safeguard more complex systems \cite{shasimplicity}.

\begin{figure}[htbp]
\centerline{\includegraphics[width=\columnwidth]{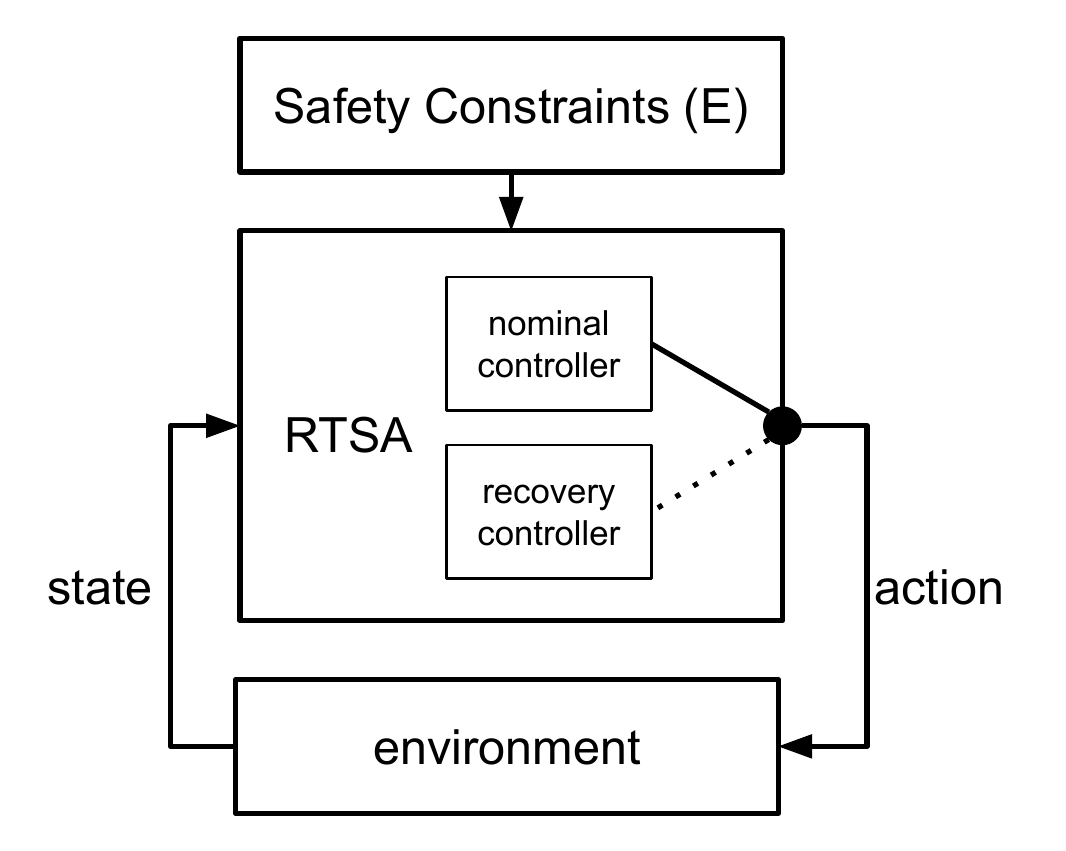}}
\caption{RTSA Scheme}
\label{diag:rtsa}
\end{figure}

RTSA ensures the safety of an autonomous agent when operating using a black-box nominal controller $\pi_n$ and switching to a simple and safe recovery controller $\pi_r$ when doing so is necessary to prevent the system from exiting a safety envelope $E$.

In order for this mechanism to work, the system needs to be able to distinguish between safe scenarios under which the operation should remain controlled by $\pi_n$ and scenarios that would likely lead to unsafe conditions in which the control should be switched to $\pi_r$.
We assume that a recovery controller $\pi_r$ is given and this work does not focus on its design or implementation. 
It is generally easier to design recovery controllers compared to nominal controllers. 
Recovery controllers are designed to be simple and safe, but are generally incapable or inefficient executing the task for which nominal controllers are designed.
For example, the experiments described in Section \ref{sec:experiments} involve an autonomous hexarotor with a recovery controller that consists of turning off thrust and deploying a parachute to minimize kinetic energy and return to the ground.

The problem that we address in this work is determining how to decide when to switch from the nominal controller $\pi_n$ to the recovery controller $\pi_r$ while balancing the trade-off between safety and efficiency. 
% maybe need to describe what pareto efficency means in this context?
This corresponds to designing a meta-controller, which consists of a policy $\pi_{RTSA}$ that switches between both controllers, as illustrated in Figure \ref{diag:rtsa}. 

There is a trade-off between safety and efficiency \cite{shasimplicity}: a very conservative $\pi_{RTSA}$ will switch to the recovery controller too often and will lead to a very safe but otherwise inefficient system. 
Conversely, $\pi_{RTSA}$ that fails to identify potentially unsafe scenarios may be efficient most of the time but may allow the system to exit the safety envelope $E$.
We postulate that the task of navigating an aircraft from an origin to a destination by following a pre-planned path is far more complex than the task of predicting whether the aircraft is operating safely within a short time horizon of a given point. 
While designing $\pi_{RTSA}$ can be challenging and remains the core of the problem, the previous notion motivates us to use simple machine learning models that are not black boxes and are both able to induce safety and allow for easy verification and interpretability.
%[The core idea is to use a safety monitor that observes the system inputs and outputs of a non-pedigreed function to verify formally-specified behaviors and states]
%[When the guard is triggered, a verified recovery control function assumes control authority to recover from an unsafe operation or state]

Successful implementation of this framework can enable the deployment of commercial off-the-shelf components or learning-enabled-components (LECs) without the need to formally verify them individually. 
Nominal controllers can be thought of as black boxes, which means that for the purposes of the RTSA system, the dynamics of the system are completely specified by the interaction of the controller with the environment and the RTSA system has no way of predicting how the controller will behave in a specific scenario.

\subsection{Runtime Safety Assurance System Design}
\label{sec:rtsa-design}
The goal of RTSA systems is to guarantee the safe operation of a system despite having black-box components as a part of its controller. 
Safe operation is specified by an envelope $E \subset S$ which corresponds to a subset of the state space $S$ within which the system is expected to operate ideally. 
RTSA continuously monitors the state of the system and switches to the recovery controller if and only if when not doing so, would lead the system to exit the safety envelope. 
In order to satisfy these considerations the following characteristics must be implemented by the RTSA system:
\begin{itemize}
	\item It must switch to the recovery control $\pi_r$ whenever the aircraft leaves the envelope.
	\item It should exhibit an efficient trade-off between safety and efficiency. To do this, it must be able to accurately predict when the aircraft is likely to leave the envelope, switching to the recovery controller only when needed.
% I realize I haven't specified what verifying means - should I?
	\item Its implementation must be easily verifiable. This means it must avoid black box models that are hard to verify such as deep neural networks (DNNs)\cite{verifsurvey}.
\end{itemize}

The RTSA system is ultimately a switch that selects between a nominal controller $\pi_n$ and a recovery controller $\pi_r$. 
The nominal controller is expected to efficiently perform the aircraft's mission but it provides no safety guarantees. 
On the other hand, the recovery controller is expected to lead to safe operational conditions.
In order to accomplish this, it is acceptable, and likely, for the recovery controller to not be very efficient at conducting the system's mission. 
%For example, a possible recovery controller could consist in reducing the vehicles' kinetic energy, in the case of aircraft this could correspond to stopping the thrust and deploying a parachute.

In this work, we demonstrate the use of terminal recovery controllers. 
These are recovery controllers that once deployed remain in control of the system until its operation terminates. 
In particular, we consider recovery controllers that act as emergency systems and minimize kinetic energy, ideally leading the vehicle to a safe stop.
An important consequence of this architecture is that it turns the RTSA into a one way switch. 
Section \ref{sec:RL} shows how to model this one-way switch. Section \ref{sec:experiments} shows how to implement and solve said model.

\section{Reinforcement Learning}
\label{sec:RL}
%- frame the problem as a sequential decision making problem
%- justify the use of RL given the black box nature of the controller, the complex dynamics (basically unknown for the RTSA) and the availability of a simulator
This section reviews the Markov decision process (MDP) framework to motivate its use in this problem and also introduces reinforcement learning (RL) as an approach for designing a switching policy for RTSA.

\subsection{Markov Decision Process Formulation}
%- briefly introduce the MDP framework
%This problem can naturally be represented as a Markov Decision Process (MDP). 
%An MDP is a mathematical framework for modeling problems involving sequential decision making under uncertainty \cite{dmu}.
%It consists of a tuple of states $S$ that represent the configuration of the environment, actions $A$ that correspond to the actions that an agent can take at each state, transitions $T$ which model how the system evolves from state $s \in S$ to state $s' \in S$ when the agent takes action $a \in A$ and rewards $R$ which correspond to rewards collected by the agent depending on each transition $(s, a, s')$.

MDPs have been used to model autonomous vehicles such as self-driving cars \cite{Bouton2020} and unmanned aircraft system \cite{Kochenderfer2015, Temizer2010}. 
In this work we use the MDP framework to model how the flight dynamics of the aircraft are affected when the RTSA system is coupled with the nominal and recovery controllers. 

We model the evolution of the flight of an aircraft equipped with an RTSA system by defining the following MDP: $\mathcal{M} = \left ( S, A, T, R \right )$ where the elements are defined below.
\begin{itemize}
	\item State space $S \in \mathbb{R}^p$: a vector representing the state of the environment and the vehicle.
	\item Action space $A \in \{\text{deploy}, \text{continue}\}$:  whether to deploy the recovery system or let the nominal controller remain in control.
	\item Transition $T(s,a)$: a function that specifies the transition probability of the next state $s'$ given that action $a$ was taken at step $s$, in our case this will be sampled from a simulator by querying $f(s,a)$.
	\item Reward $R(s,a,s')$, the reward collected corresponding to the transition. This will be designed to induce the desired behavior.
\end{itemize}

In this model, the RTSA system is considered the agent while the states correspond to the position, velocity and other relevant information about the aircraft with respect to the envelope and the actions correspond to deploying the recovery controller or not.
The agent receives a large negative reward for abandoning the envelope and a smaller negative reward for deploying the recovery controller in situations where it was not required. 
This reward structure is designed to heavily penalize situations in which the aircraft exits the safety envelope and simultaneously disincentivize unnecessary deployments of the recovery controller.
 The rewards at each step are weighted by a discount factor $\gamma < 1$ such that present rewards are worth more than future ones. 
 
A policy $\pi(s)$ is a function that determines which action $a$ to perform at each state $s$. 
The utility of following policy $\pi$ from state $s$ is denoted $U^\pi(s)$ and is typically referred to as the value function. Solving an MDP corresponds to crafting a policy $\pi^*$, that leads to maximizing the expected utility $\mathbb{E} \left [ U^\pi \right ]$:

\begin{align}
	\pi^*(s) = \arg \max_\pi U^\pi (s)
\end{align}

Simple cases of MDPs such as those with discrete and small state and action spaces where the transition dynamics are known and have a closed form can be solved exactly using techniques derived from dynamic programming such as the policy iteration algorithm \cite{dmu}.

However, in the case of autonomous flight analyzed here, the state space is high dimensional because it includes all the variables needed to model realistic flight dynamics, avionics, sensor readings and environmental conditions such as wind, the transition model is also highly complex and relies on numerical methods to compute the evolution of a flight.
% don't like this transition

Additionally, we consider the nominal controller to be a black box. All of these conditions combined lead us to operate under the assumption that the transition function is unknown and we do not have access to it. 
We do, however, have access to simulators from which we can query experience tuples $(s,a,r,s')$ by providing a state $s$ and an action $a$ and fetching the next state $s'$ and associated reward $r$. 
In this setting we can learn a policy from experience with RL.

\subsection{Reinforcement Learning}

Our agent needs to learn to choose actions that maximize its long-term accumulation of rewards by observing the outcomes of its actions in the form of state transitions and rewards. 
This type of problem comes with many challenges, among them the fundamental trade-off between exploration and exploitation, which refers to the balance between exploring new possibilities and exploiting already acquired knowledge.
RL has successfully been applied to many fields \cite{dqn, julianfire, Menda2019}. 

%The variety of approaches to solve RL problems can be grouped in many ways. 
%A first division corresponds to algorithms that try to explicitly infer the MDP and later solve it using traditional MDP algorithms which are referred to as model-based methods. 
%Alternatively, model-free methods are those that do not explicitly reconstruct the MDP and craft a policy in a somewhat roundabout way. 
%Many of the model-free methods construct a policy by fitting a function typically referred to as the 
One approach to RL involves constructing a $Q$-function that estimates the value of performing each possible action at each possible state. The $Q$-function has the following structure: $Q: S \times A \rightarrow \mathbb{R}$, and represents the quality of a state-action combination:
\begin{align}
	Q(s,a) &= R(s,a) + \gamma \sum_{s'} T( s' \mid s,a) U(s')
\end{align}

When the MDP is known, the $Q$-function can be computed using dynamic programming as shown in the Bellman Equation below:

\begin{align}
\label{eq:qbell}
	Q(s,a) = R(s,a) + \gamma \sum_{s'} T(s'\mid s,a) \max_{a'} Q(s',a')
\end{align}

Optimal policies can be extracted after performing the above computation and taking an action
\begin{align}
a^* = \arg \max_a Q(s,a)
\end{align}
that maximizes the $Q$-function for a given state.

The $Q$-function can also be estimated using simulators or historic data as in the popular $Q$-learning algorithm \cite{Watkins92q-learning}. 
$Q$-learning is a model-free reinforcement learning algorithm, and involves applying incremental updates to estimations of the Bellman Equation (\ref{eq:qbell}):
\begin{align}
	Q(s,a) &= R(s,a) + \gamma \sum_{s'} T( s' \mid s,a) U(s')\\
	&= R(s,a) + \gamma \sum_{s'} T( s' \mid s,a) \max_a Q(s',a')
\end{align}
The key observation is that instead of using the transition function $T$ and the reward function $R$, we use the observed state $s$, next state $s'$ and reward $r$ obtained after performing action $a$. The algorithm is outlined below.

% TO DO: switch to algorithm format
% Option 1 (comes with the template)

%\begin{algorithmic}[1]
%	\STATE $t \Leftarrow 0$
%	\STATE $s_0 \Leftarrow$ initial state
%	\STATE $Q \Leftarrow$ initialize
%	\FOR{\texttt{every iteration}}
%	\STATE Choose action $a_t$ \COMMENT{based on some exploration strategy}
%	\STATE Observe new state $s_{t+1}$ and reward $r_t$
%	\STATE $Q(s_t,a_t) \leftarrow Q(s_t, a_t) + \alpha(r_t + \gamma \max_a Q(s_{t+1}, a) - Q(s_t, a_t))$
%	\STATE $t \Leftarrow t+1$
%	\ENDFOR
%\end{algorithmic}

\iffalse
% Option 2 (looks way better IMO)
\begin{algorithm}
  \caption{Q-Learning}\label{alg:q-learning}
  \begin{algorithmic}[1]
    \Procedure{Q-learning}{$T,R$}
      \State $t \gets 0$
      \State $s_0 \gets$ initial state
	  \State $Q \gets$ initial values
      %\State $r\gets a\bmod b$
      \For{\texttt{every step}}
		\State Choose action $a_t$ based on Q \Comment{and an exploration strategy}
        \State Observe new state $s_{t+1} =T(s_t, a_t) $ and reward $r_t = R(s,a,s_{t+1})$
        \State $Q(s_t,a_t) \leftarrow Q(s_t, a_t) + \alpha(r_t + \gamma \max_a Q(s_{t+1}, a) - Q(s_t, a_t))$
        \State $t \leftarrow t+1$
      \EndFor
    \EndProcedure
  \end{algorithmic}
\end{algorithm}
\fi
% english description of q-learning instead of above
\subsubsection{$Q$-Learning}
\label{sec:qlearning}
$Q$-learning consists of initializing the values in the $Q$-table randomly and then at step $t$ with state $s_t$, select an action $a_t$ based on the table and an exploration strategy.
%\chris{would like to discuss this more}
Then, upon observing the a new state and reward
\begin{align}
	s_{t+1} \sim T(\cdot \mid s_t, a_t), && r_t = R(s,a_t)
\end{align}
the table is updated as follows:
\begin{align}
	Q(s_t,a_t) = Q(s_t, a_t) + \alpha (r_t + \gamma \max_a Q(s_{t+1}, a) - Q(s_t, a_t))
\end{align}
where $\alpha$ is the learning rate.

$Q$-learning as described above enables the estimation of tabular $Q$-functions which are useful for small discrete problems. 
However, cyber-physical systems often operate in contexts which are better described with a continuous state space. 
The problem with applying tabular $Q$-learning to these larger state spaces is not only that they would require a large state-action table but also that a vast amount of experience would be required to accurately estimate the values. 
An alternative approach to handle continuous state spaces is to use $Q$-function approximation where the state-action value function is approximated which enables the agent to generalize from limited experience to states that have not been visited before and additionally avoids storing a gigantic table. 
Policies based on value function approximation have been successfully demonstrated in the aerospace domain before \cite{julian2016policy}.
 
There exist a variety of approaches to approximate the $Q$-function. Some rely on local information and use a notion of distance to interpolate between previously visited states, while others perform global approximation. 
DNNs have been used successfully as value function approximators \cite{dqn, alphago}.

Learning a policy with value function approximation consists of specifying a family of functions that depend on parameters and then finding the parameters that approximate the optimal $Q$-function by interacting with the environment. 
This idea has made it possible to use state-action value function algorithms in higher dimensional problems and even problems with continuous states such as ours.
After its introduction in \cite{dqn} for RL tasks, a common approach is to use DNNs as function approximators.
 
 Although DNN function approximators have demonstrated impressive performance in various domains \cite{dqn, alphago, Bouton2020, julianfire}, they have a fundamental drawback in our setting: they are considered black boxes.
 Given that it is difficult to formally verify the behavior of DNNs \cite{liu2019algorithms}, we are unlikely able to use them to implement the switching mechanism in the RTSA architecture.
Instead, we will restrict our attention to linear value function approximation, which involves defining a set of features $\Phi(s,a) \in \mathbb{R}^m $ that captures relevant information about the state and then use these features to estimate $Q$ by linearly combining them with weights $\theta \in \mathbb{R}^{m \times \mid A \mid}$ to estimate the value of each action. In this context, the value function is represented as follows:
 
\begin{align}
\label{eq:linq}
	Q(s,a) = \sum_{i=1}^m \theta_i \phi_i(s, a) =  \theta^T \Phi(s,a)
\end{align}

Our learning problem is therefore reduced to estimating the parameters $\theta$ and selecting our features.
Typically, domain knowledge will be leveraged to craft meaningful features $\Phi(s,a) = (\phi_1(s,a), \phi_2(s,a),...,\phi_m(s,a))$, and ideally they would capture some of the geometric information relevant for the problem, e.g. in our setting, heading, velocity and distance to the geofence.

The ideas behind the $Q$-learning algorithm can be extended to the linear value function approximation setting. 
Here, we initialize our parameters $\theta$ and update them at each transition to reduce the error between the predicted value and the observed reward. 
The algorithm is outlined below and it forms the basis of the learning procedure used in the experiments in Section \ref{sec:experiments}.
\iffalse
\begin{algorithm}
  \caption{Linear Approximation Q-Learning}
  \label{alg:linear-q-learning}
  \begin{algorithmic}[1]
    \Procedure{LinearApprox-Q-Learning}{$T,R$}
      \State $t \gets 0$
      \State $s_0 \gets$ initial state
	  \State $\theta \gets$ initial values
      %\State $r\gets a\bmod b$
      \For{\texttt{every step}}
		\State Choose action $a_t$ based on $\theta_a^T \Phi(s)$ \Comment{and an exploration strategy}
        \State Observe new state $s_{t+1} =T(s_t, a_t) $ and reward $r_t = R(s,a,s_{t+1})$ 
        \State $\Theta \gets \Theta + \alpha  (r_t + \gamma \max_a \Theta^T \Phi(s_{t+1}) -$
          $\Theta^T \Phi(s_t) ) \Phi(s_t)$
        \State $t \leftarrow t+1$
      \EndFor
    \EndProcedure
  \end{algorithmic}
\end{algorithm}
\fi
\subsubsection{Linear Approximation Q-Learning}
\label{sec:linqlearning}
Similar to $Q$-learning, described earlier, Linear Approximation $Q$-learning begins by initializing the parameters $\theta$, usually randomly. At step $t$ with state $s_t$, select an action $a_t$ based on $\theta^T \Phi(s,a)$  and an exploration strategy.
Then, upon observing the a new state and reward
\begin{align}
	s_{t+1} \sim T(\cdot \mid s_t, a_t), && r_t = R(s,a_t, s_{t+1})
\end{align}
the parameters are updated as follows:
%\chris{if I parameterize Phi with (s,a) then what happens to the term $\Phi(s_{t+1},a)$ in the linear Q-learning update below? a bit confused now}
\begin{align}
	\theta = \theta + \alpha  (r_t + \gamma \max_a \theta^T \Phi(s_{t+1}, a) - \theta^T \Phi(s_t, a_t) ) \Phi(s_t, a_t)
\end{align}

Notice that at each step the action $a$ is selected according to the estimated $Q$-function and an exploration strategy. 
The exploration strategy is designed to guarantee that the algorithm converges to an optimal policy and usually consists of random exploration. 
In our setting, the exploration strategy requires careful tuning to avoid consistently deploying the recovery controller during the learning stage.

\subsection{RL for RTSA}
%Given the MDP formulation of RTSA and the learning algorithms outlined above, we now delve deeper into the technical details of how to utilize RL to learn a policy to implement an RTSA system.
In our context where we have a nominal controller $\pi_n$ and an emergency controller $\pi_r$, the goal is to learn an optimal policy $\pi_{RTSA}$ that switches control from $\pi_n$ to $\pi_r$ when needed to guarantee safety. 
It follows that the controller of the system $\pi$ has the following form:

\begin{align}
	\pi(s) = \begin{cases}
\pi_n(s) & \text{ if } \pi_{RTSA}(s) = \text{continue} \\ 
\pi_r(s)& \text{ if } \pi_{RTSA}(s) = \text{deploy} 
	\end{cases}
\end{align}

%-- transition - coupling of the controller or rcf and the dynamics
Additionally, both $\pi_n$ and $\pi_r$ are black box controllers which means that we have no model for them and can only query them in a generative manner. 
In this case, each step produced by the simulator corresponds to sampling $T$ which is determined by the composition of $\pi$ and $f(s,a)$, the step function of the simulator:

\begin{align}
	s_{t+1} &\sim T\left (\cdot \mid s_t,a_t \right ) = f \left ( s_t,a_t \right ) =\\ f\left (s_t, \pi \left ( s_t,a_t \right )  \right ) &= \begin{cases}
f\left (s_t,  \pi_n(s_t) \right ) & \text{ if } \pi_{RTSA}(s_t) = \text{continue} \\
f \left ( s_t,  \pi_r(s_t)\right )& \text{ if } \pi_{RTSA}(s_t) = \text{deploy} 
	\end{cases}
\end{align}
Here, we assume $f$ encompasses all the flight dynamics and returns the next state of the system in either the nominal or emergency regime depending on the value of $\pi_{RTSA}(s)$.

%-- reward - crafted to serve as a dial between safety and efficiency
Finally, we define the reward function $R$ such that maximizing the expected utility induces the desired behavior outlined in Section \ref{sec:rtsa-design} as follows:

\begin{align}
	R\left ( s,a, s' \right ) = \begin{cases}
	 0 & s' \in E\\ 
 -\alpha & \text{if}\ a = \text{deploy}  \\ 
-1 & \text{if}\ s' \notin E
\end{cases}
\end{align}

There is no cost associated with letting the aircraft operate under safe conditions, a cost of $\alpha$ associated to engaging the emergency system and a fixed negative cost associated with exiting the safety envelope. 
We can control the trade-off between safety and efficiency by assigning different value for $\alpha$.
 Large values of $\alpha$ will discourage the deployment of $\pi_r$. 
 This allows us to tweak the behavior of the meta-controller and later analyze whether it consistently dominates a baseline meta-controller. 

%- policy - intelligible policy to enable the verfication of the RTSA so that the system is trutworthy
\subsection{Policy Design and Feature Crafting}
%As described in Section \ref{sec:rtsa-design}, the implementation of $\pi_{RTSA}$ has to be transparent so that its behavior can be formally verified. 
%This means that DNNs or other highly complex models cannot be used for value function approximation.
%Only models that can be easily verified, and ideally are intelligible by human experts, can be used. 
We restrict our function family to linear functions which can be easily understood and verified. 
A major drawback, however, is that linear functions are less expressive than DNNs, which makes their training more difficult and requires careful crafting of features. 

In the aerospace domain, a high fidelity representation of the physical world requires taking into account many magnitudes which leads to a high dimensional state representation.
However, linear value function approximation as defined in Equation \ref{eq:linq} operates over features $\Phi(s,a)$. 
This means that we do not need to use all the information in a state and can instead extract a few features that capture the relevant information that is needed to determine whether the vehicle is likely to exit the envelope or not.
 For our experiments, we selected features that correspond to the velocity vector of the aircraft, a vector that represents the direction and distance of closest approach to the edge of the geofence, a vector that represents the wind conditions and an indicator variable that represents whether the recovery system has been deployed or not. 
 Feature crafting will be further discussed in Section \ref{sec:experiments}.

Despite the relative simplicity of linear value function approximators when compared to DNNs, we observed that they are able to capture relevant information about the environment and are well suited for this task. 
Recall that action $a$ is selected at state $s$ if it maximizes $\theta^T \Phi(s,a)$. Features should be crafted considering the situational information they provide. 
\iffalse
This observation is supported by analyzing the geometric implications of linearly combining features. Ignoring the exploration strategy for the moment, at each step our agent will perform the action 
\begin{align}
	a &= \arg \max_a Q(s,a)\\
	&= \arg \max_a \theta^T \Phi(s,a)
\end{align}
which can be rewritten as
\begin{align}
a &= \arg \max_a Q(s,a) = \left\{\begin{matrix}
\sum_{i=1}^m \theta_{i,1}(a) \phi_i(s) & \text{ if }a= \text{ false} \\
 \sum_{i=1}^m \theta_{i,0}(a) \phi_i(s)& \text{ if }a= \text{ true}
\end{matrix}\right.\\
&= \arg \max_a \left \{ \sum_{i=1}^m \theta_{i,0} \phi_i(s), \sum_{i=1}^m \theta_{i,1} \phi_i(s) \right \}\\
&= \arg \max_a \left \{ \Theta_0^T \Phi(s), \Theta_1^T \Phi(s) \right \}.
\end{align}
It is easy to observe that the action $a$ taken at state $s$ is the one that maximizes the dot product between the feature vector $\Phi(s)$ and the parameter vector $\theta$. 
Dot products are associated to the angle between vectors so we can conclude that the action selected corresponds to the one whose parameter vector is closest aligned with the feature vector. 
This geometric property informs how our features should be crafted to design a useful policy. 
\fi
For example, if some of the features represent the velocity vector of the autonomous aircraft and others represent the direction of closest approach to the edge of the geofence, then if both vectors are aligned, the vehicle is heading in a direction in which it is likely to exit the geofence.
Therefore, we would expect the policy to learn negative weights for $a=$ continue, which is to select $\pi_n$ and positive weights for $a=$ deploy, which selects $\pi_r$. 

%Accordingly, features should be selected such that there exists a separating hyperplane between the values of the features for which the recovery controller should be deployed and the values for which the nominal controller should retain control.

Given our definition of the reward function $R$, the linear approximation of the value function should be able to represent the significant change in reward that occurs when $\pi_r$ is deployed. 
A way to facilitate this is to append an extra feature that is $1$ if $\pi_r$ has been deployed and $0$ otherwise.
%\begin{align}
%	\phi_* = \begin{cases}
%1 & \text{ if } \pi_r \text{ has been deployed} \\ 
%0 & \text{ otherwise} 
%	\end{cases}
%\end{align}
This feature enables the linear approximator to shift its estimation of the value of state-action tuples once $\pi_r$ is deployed by a learnable constant corresponding to the coefficient associated with this feature.
% CHRIS AQUI AQUI AQUI AQUI

\subsection{Exploration}
%As stated earlier, we set out to develop an approach that leads to the design of a meta-controller $\pi_{RTSA}$ for high dimensional problems that may involve non-linear dynamics and a black-box nominal controller $\pi_n$. 
%This restricted our selection of learning algorithms to model-free methods and necessitates relying on value function approximation given the high-dimensionality and continuous nature of the aerospace domain discussed in Section \ref{sec:RL}.
%Additionally, given the transparency and verifiability requirements of the RTSA system discussed in Section \ref{sec:rtsa-design}, we further restrict our approximator function class to linear functions.

%Recall that our policy depends on our estimation of the $Q$-function and we are using linear value function approximation to do so. After the features are specified for each state, the only problem remaining is to learn the parameters by which to linearly combine them to estimate the value function for each action.
%That is, the goal is then to learn a set of parameters $\theta$ such that $Q(s,a) = \theta^T \Phi(s,a)$ estimates the real $Q$-function for any combination of states and actions. 

%A natural approach to estimate the parameters in a model-free approach with a generative framework as ours would be to do Linear Approximation $Q$-learning \cite{dmu} as described in Section \ref{sec:linqlearning}.

%A likely problem with this approach, which we encountered in our experiments and is further explained in Section \ref{sec:experiments}, is the exploration strategy.
An exploration strategy is critical for balancing the exploration versus exploitation trade-off mentioned in Section \ref{sec:RL}, and it typically encompasses some sort of random action selection. Often, the $\epsilon$-greedy strategy is used, where at each step during training with probability $0<\epsilon<1$ we select an action randomly and with probability $1-\epsilon$ we select the greedy action prescribed by the policy we are training. 
In our setting, this exploration strategy is not viable because deploying the recovery controller is a non-reversible terminal action.
Once the recovery controller is deployed, it remains in control of the aircraft until the episode terminates, which biases the state and action tuples observed in a trajectory and prevents the learning algorithm from capturing the reward signal associated with successfully finishing a mission. 
A naive solution to this problem would be to use a low value of $\epsilon$, but the problem remains as the exploration strategy is queried at every step in the simulation and a single episode is composed of hundreds or thousands of steps, making the probability of randomly deploying the recovery controller during one episode very high.

One approach to address this problem is to avoid or reduce the chance of random exploration. 
We dramatically increase the likelihood of observing episodes where the mission is completed successfully without exiting the envelope, but we also bias the learning process towards exploitation. 
It is well known that in order for a policy to converge under $Q$-learning, exploration must proceed indefinitely \cite{gliepaper}.
%This means that if in the long run a state is visited infinitely often, then each feasible action in that state has to be chosen infinitely often.
Additionally, in the limit of the the number of steps, the learning policy has to be greedy with respect to the $Q$-function \cite{gliepaper}.
Accordingly, avoiding or dramatically reducing random exploration can negatively affect the learning process and should be avoided.

%\textcolor{blue}{A viable solution to this problem is to turn the policy into a stochastic policy by adding a softmax operator on top of the value function approximation. 
%This means that the policy itself will be randomly sampled but the distribution of the actions depends on the estimation of the values of each action.}
%This approach, coupled with parameter initialization in which deployment of the recovery controller is disincentivized, was shown to be a viable approach in Section \ref{sec:experiments}. 
%As training progresses, we \textcolor{blue}{decrease the temperature} which is analogous to being greedy in the long term, to satisfy the second requirement for convergence.

Ideally, we want the learned policy to have a low probability of falsely deploying the recovery controller to satisfy the specifications in Section \ref{sec:rtsa-design}. 
This turns out to be important in the learning process too because a high rate of false deployment significantly reduces the amount of experience collected by the agent where the mission was completed. The agent must experience completing the mission many times in order to capture the positive reward signal associated with completing safe missions without interruption.
 This can be achieved with a careful initialization of the model weights $\theta$ that reduces the value of deploying the recovery controller. 
 However, determining how to do this specifically and the magnitude by which it should be reduced is a complex problem itself.
 Instead, we rely on RL to solve this issue.
 
 Instead of randomly initializing the parameters of the $Q$-function approximation and then manually biasing the weights to decrease the chance of randomly deploying the recovery controller, we can use a baseline policy to generate episodes in which the RTSA system exhibited a somewhat acceptable performance. 
 From these episodes, we learn the parameters in an offline approach known as batch reinforcement learning. It is only after we learn a good initialization of our parameters that we then start the training process of our policy $\pi_{RTSA}$.

For this purpose and to have a benchmark to compare our approach to, we define a baseline policy that consists of shrinking the safety envelope by specifying a distance threshold $\delta > 0$. 
When the vehicle reaches a state that is less than $\delta$ distance away from exiting the envelope, the recovery controller is deployed. 
This naive approach serves both as a baseline for our experiments and also provides us with experience to initialize the weights of our policy before we do on-policy learning.

% CHECKPOINT LUNES

\section{Experiments}
\label{sec:experiments}

To demonstrate our approach we implemented an RTSA system as described in Section \ref{sec:rtsa-design} by training a policy using the techniques described in Section \ref{sec:RL}. 
% TO-DO get PX4 autopilot ref
The goal is to demonstrate the viability of the approach as many civil unmanned aircraft applications are tending towards the use of open-source autopilots such as the PX4, which is an open-source autopilot system, for their relatively low cost of implementation and applicability across a wide range of vehicle types.

For all our experiments we used a simulator designed for Small Unmanned Aircraft System 
% TO-DO: figure out simulator ref
(sUAS) designed as a part of the UAS traffic management system project \cite{ge-suas}.
%\chris{confirm with Jerry if this is the right ref}
This software package is intended to simulate operations of small unmanned aircraft systems weighing less than 55 lb. 
Unmanned aircraft of this size have been flown as model aircraft for recreation and sports uses for many decades. 
Wide-spread use of sUAS for non-recreational purposes has been limited, particularly for applications requiring operations beyond the visual line of sight (BVLOS) from the operator \cite{ge-suas}.
%\chris{confirm with Jerry if this is the right ref}
However, driven by the potential societal and economic benefits that can be generated from the use of sUAS, new systems such as RTSA can enable civilian low-altitude airspace unmanned aircraft operations, particularly sUAS BVLOS operations by providing airspace corridors or geofences. 
%This situation is a natural example for the use of RTSA as the safety envelope described in Section $\ref{sec:rtsa}$ can easily represent a geofence.

Our experiments used the Small Unmanned Aircraft Trajectory Modeler (SAT), which is a suite of MATLAB simulation tools. 
It is designed to support trajectory patterns for diverse airframe configurations, including horizontal take-off and landing, vertical flight, and hybrid vertical take-off and landing aircraft, both under normal conditions and under a variety of realistic potential hazards, including adverse environmental conditions such as strong winds. 

\begin{figure}[htbp]
\centerline{\includegraphics[width=\columnwidth]{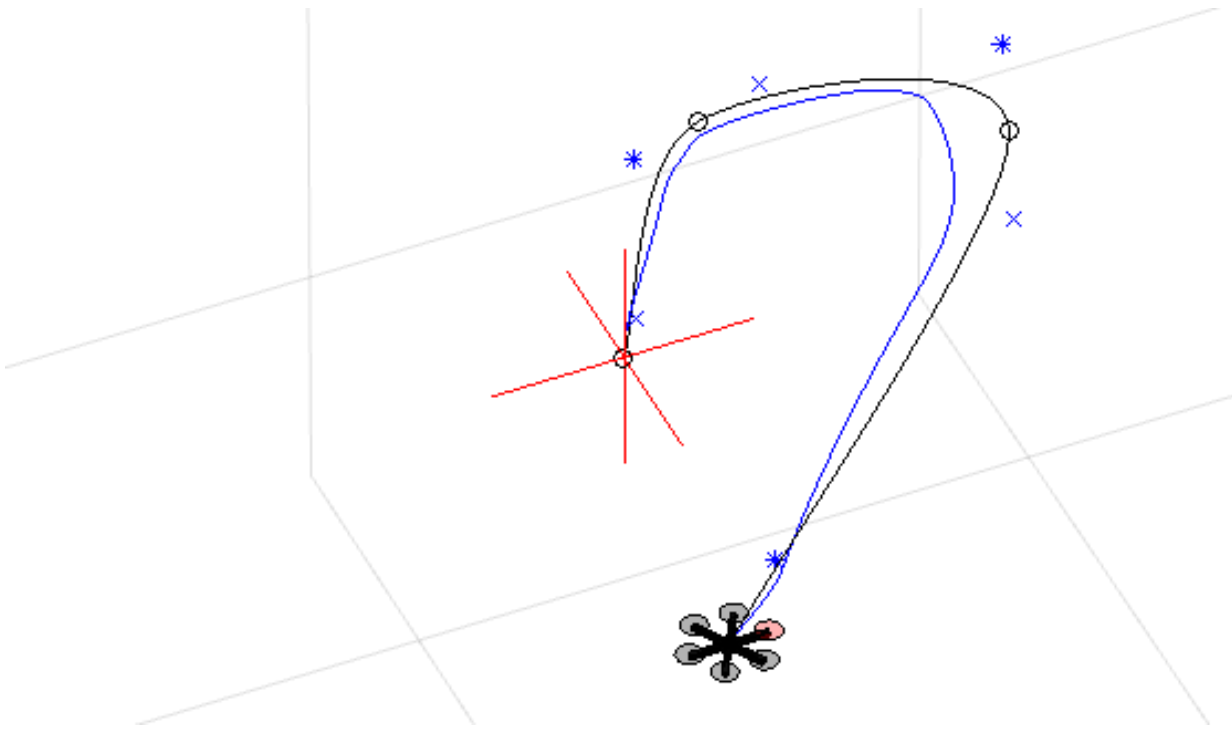}}
\caption{Environment}
\label{fig:3denv}
\end{figure}

We used configuration composed of a hexarotor simulator that models lift rotor aircraft features and includes a three dimensional Bezier curve trajectory definition module, nonlinear multi-rotor dynamics, input/output linearization, nested saturation, a cascade PID flight control model, and an extended Kalman filter estimation model. 
An illustration of the simulator environment in three dimensions is included in Figure \ref{fig:3denv} \cite{ge-simulator}.
%\chris{confirm ref with jerry}

An RTSA system is composed of a nominal controller and a recovery controller and a safety envelope which we describe for our experiments below.

% TO-DO: explain how the matlab autopilotworks

\begin{itemize}
	\item Nominal controller $\pi_n$: The nominal controller is an autopilot that implements a path following controller intended to represent an open-source off the shelf solution.
	\item Recovery controller $\pi_r$: The emergency controller corresponds to turning off the rotors and deploying a parachute which was modeled using a simplified model that introduces a drag coefficient that only affects the $z$ coordinates in the simulation.
	\item Safety envelope $E$: In this setting the safety envelope $E$ corresponds to a three dimensional hyper-rectangle that completely covers the planned trajectory for the mission along with extra volume that would account for the airspace corridor assigned for the mission.
\end{itemize}

\begin{figure}[htbp]
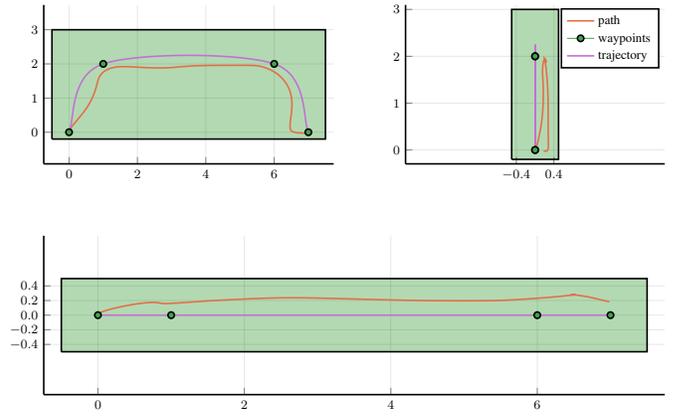

\begin{center}
\includestandalone[width=\columnwidth]{img/3d-traj_plot}
\end{center}
\caption{Example of environment configuration and episode data with wind.}
\label{fig:example_env}
\end{figure}

%\begin{figure}[htbp]
%\centerline{\includegraphics[width=.45\textwidth]{img/traj-example}}
%\caption{Example of Environment Configuration}
%\label{fig:example_env}
%\end{figure}

An example configuration is illustrated in Figure \ref{fig:example_env}. 
Where each of the subplots correspond to a different plane of the three dimensional space. 
The green box corresponds to the geofence $E$, the green dots to the waypoints that define the path represented by the purple line and the red line corresponds to the trajectory followed by the drone during one simulation with random wind. 
The aircraft has to remain within the green three dimensional hyper-rectangle as it completes its mission which is to fly from the first waypoint located at the origin to the last one at the end of the path.
\subsection{Environment}
\label{sec:environment}

\begin{figure}[htbp]
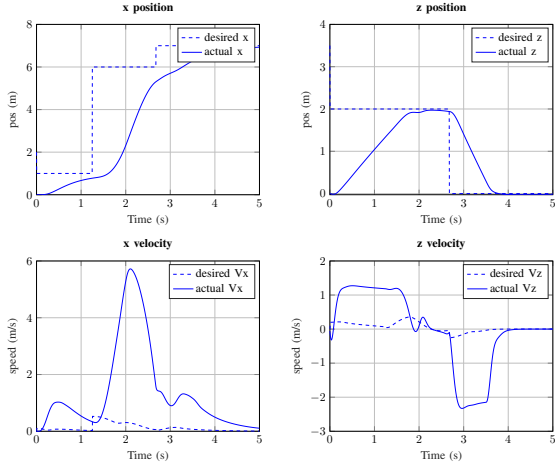

\begin{center}
\includestandalone[width=\columnwidth]{tik2}
\end{center}
\caption{Example of simulation episode location and velocity data.}
\label{fig:sample-data}
\end{figure}

The state space in our simulation is comprised of more than $250$ variables. 
Some correspond to simulation parameters such as sampling time, simulation time and physical constants.
Another set of variables represent physical magnitudes such as velocity scaling factors, the mass of components of the hexarotor, distribution of the physical components of the hexarotor, moments of inertia and drag coefficients. 
Other variables represent maximum and minimum roll, pitch and yaw rates, rotor speed and thrust.
The sensor readings, their biases, frequencies and other characteristics are also represented by other variables. 
Other variables represent the state of the controller and the actions it prescribes for the different actuators in the vehicle.
And a few of them correspond to the position and velocity of the hexarotor during the simulation. Figure \ref{fig:sample-data} shows the evolution of the position and velocity variables for a simulation episode corresponding to the example environment configuration.

All of these variables are needed to completely specify the state of the world in the simulation and illustrate the high dimensional requirements of a somewhat accurate representation of flight dynamics in a simulator. 
In principle, all these variables would be needed to specify a policy for our RTSA system, but as discussed in Section \ref{sec:RL} we can rely on value function approximation by crafting a set of informative features which significantly reduces the dimensionality of our problem.

%\begin{center}
%\setlength{\figW}{4} % when added inside the  center environment it has no effect outside it
%\input{../matlab2t/plotall}
%\end{center}

%\begin{figure}[htbp]
%\centerline{\includegraphics[width=.45\textwidth]{img/environ2-placeholder}}
%\caption{Example of simulation episode location and velocity data}
%\label{fig:sample-data}
%\end{figure}

The simulator exhibits a deterministic behavior: given all the values of the simulation variables the next state is fully specified given the action of the agent.
However, in order to account for unforeseen circumstances that the system could encounter in real operational conditions and to showcase the robustness of our approach, we infused stochasticity to the environment by randomly generating a wind vector field for each simulation episode.
% win params: normal, mu = 0, sigma_mult = 2.5
We specified the distribution of the wind vector field such that without RTSA the aircraft exits the envelope about a quarter of the time.

\subsection{Baseline}
As described in Section \ref{sec:rtsa}, safeguard systems could be designed by human engineers and a natural approach would be to craft a switch that depends on the distance of the aircraft to the edge of the geofence. 
A threshold $\delta>0$ is specified and if the aircraft's distance to the edge of the geofence is $\delta$ or less, the the recovery controller is deployed:

\begin{align}
	\pi_{B}(s, \delta) = \begin{cases}		
0 & \text{ if } d(s,E) \leq \delta \\ 
1 & \text{ if } d(s,E) > \delta
\end{cases}
\end{align}

The baseline policy $\pi_B$ is formally described above, where $d(s,E)$ is the distance of closest approach between the vehicle and the edge of geofence. 
We explicitly parameterized the policy by $\delta$ because it serves as a tuning mechanism between safety and efficiency and we will evaluate the performance of $\pi_B$ for different values of $\delta$ in Section \ref{sec:exp-results} and compare it to the learned policy. This baseline policy corresponds to shrinking the envenlop $E$ in every direction and deploying $\pi_r$ when the vehicle exits this smaller region.

\begin{figure}[htbp]
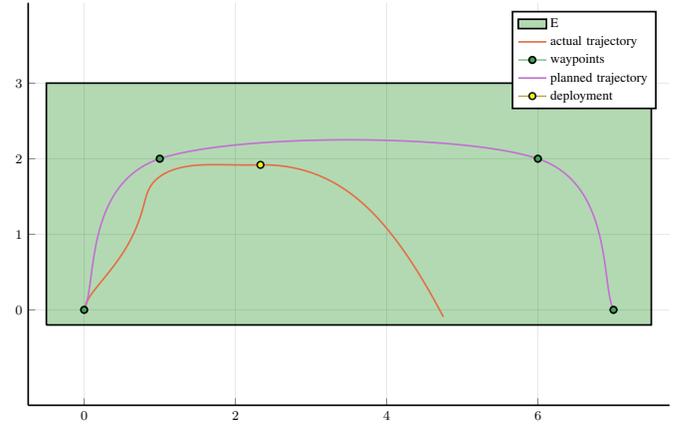

\begin{center}
\includestandalone[width=\columnwidth]{img/traj_plot}
\end{center}
\caption{Example of simulation episode location and velocity data along the $x-y$ plane.}
\label{fig:mission}
\end{figure}

For demonstration purposes and to keep the computational requirements tractable we designed a simple mission by specifying four waypoints that correspond to the initial position of the hexarotor, two other waypoints and final destination on the ground. 
This configuration is displayed in Figure \ref{fig:mission} along with the trajectory planned with by the trajectory definition module along the $x-y$ plane.

%\begin{figure}[htbp]
%\centerline{\includegraphics[width=.45\textwidth]{img/mission-placeholder}}
%\caption{Mission definition along the X-Z plane}
%\label{fig:mission}
%\end{figure}

\subsection{Training}
We proceed as described in Section \ref{sec:RL}. 
Recall that the objective of the training procedure is finding the optimal values of $\theta$ which are the weights that specify the linear value function approximation of $Q$. 
There is one entry in $\theta$ associated to each feature and each action.

\begin{figure}[htbp]
\centerline{\includegraphics[width=\columnwidth]{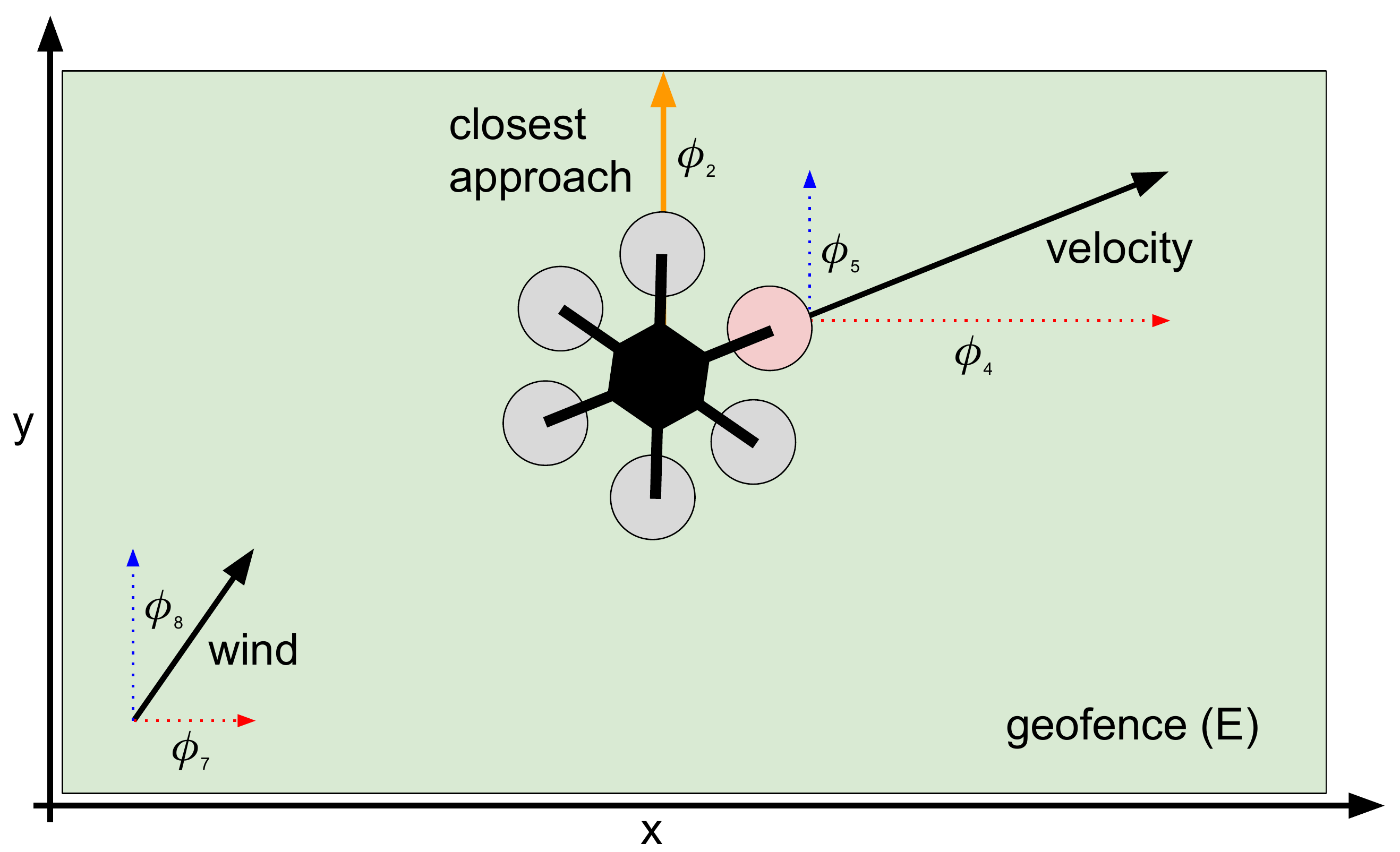}}
\caption{Features}
\label{fig:features}
\end{figure}

% features etc
The following features were used for these experiments (see Figure \ref{fig:features} for an illustration in the $x-y$ plane):
\begin{itemize}
	\item $\phi_1$: distance to the edge of the geofence in component $x$
	\item $\phi_2$: distance to the edge of the geofence in component $y$
	\item $\phi_3$: distance to the edge of the geofence in component $z$
	\item $\phi_4$: speed in component $x$
	\item $\phi_5$: speed in component $y$
	\item $\phi_6$: speed in component $z$
	\item $\phi_7$: wind speed in component $x$
	\item $\phi_8$: wind speed in component $y$
	\item $\phi_*$: indicator of deployment
\end{itemize}

We chose these features because, as described in Section \ref{sec:RL}, the linear value function approximator is able to capture the relationship about the angle between vectors. 
The angles between the velocity vector, vector of closest approach to geofence and wind vector should, intuitively, play a prominent role in the switching behavior of $\pi_{RTSA}$.

% Reward function
The reward function was crafted to induce the desired behaviors established in \ref{sec:rtsa-design}:

\begin{align}
	R\left ( s,a, s' \right ) = 
\begin{cases}
 0 & s' \in G \\ 
 -\alpha & \text{if}\ a = \text{true}  \\ 
-1 & \text{if}\ s' \notin G
	\end{cases}
\end{align}

With this definition of $R$ at hand, and leaving everything else fixed, we can tune the trade-off between safety and efficiency by specifying different values for $\alpha$. 
Large values of $\alpha$ deincentivize the deployment of $\pi_r$ and small values have the converse effect. 
Decreasing the rate at which $\pi_r$ is deployed reduces the amount of false positive outcomes and potentially allows for more efficient performance in terms of the times that the mission is completed uninterrupted, at the expense of safety.
Incentivizing the deployment of the $\pi_r$ leads to a safer behavior but can ultimately make the system very inefficient if the recovery controller is deployed too often.

% linear q learning, issues with deployment, batch learning
For our initial experiments, the parameters were randomly initialized and we used an $\epsilon$-greedy exploration strategy.
However, we observed the issues described in Section \ref{sec:RL} where the recovery controller was deployed in all episodes.
After trying different values of $\epsilon$ and still observing the same behavior, we concluded that random initialization of $theta$ could highly likely explain this situation. 

As a sanity test, we ran some episodes where the initial weights where highly biased to deincentivize the deployment $\pi_r$ by manually modifying the value of the affine term in the model. 
We observed that episodes were terminating successfully without consistently deploying $\pi_r$.

The initial values of $\theta$ were, however, arbitrary and so we proceeded to soft-start the training by using episodes generated with the baseline policy to estimate the parameters. We then used conventional linear approximation $Q$-learning as described in Section \ref{sec:linqlearning}.
We performed some minor hyper-parameter tuning to select a learning rate.

%Figure \ref{fig:baseline-geo} illustrates the subregion of $E$ where the baseline policy allows $\pi_n$ to remain and control and Figure \ref{fig:baseline-geo} is analogous por the learned policy.

%\begin{figure}[htbp]
%\centerline{\includegraphics[width=.45\textwidth]{img/example-policy}}
%\caption{Baseline Policy Plot}
%\label{fig:baseline-geo}
%\end{figure}

%\begin{figure}[htbp]
%\centerline{\includegraphics[width=.45\textwidth]{img/example-policy}}
%\caption{Linear Policy Plot}
%\label{img:linear-geo}
%\end{figure}

%Notice that in both cases $\pi_r$ would be deployed if the vehicle exits $E$ as the operational region is a subset of the geofence which is consistent with the requirements established in Section \ref{sec:rtsa-design}.

\subsection{Results}
\label{sec:exp-results}
In order to assess the performance of the policy, we ran multiple episodes with random wind conditions. We used the same random seed for all the policies described below and compared their results.
\begin{itemize}
	\item $\pi_n$: the nominal controller with no safety assurance.
	\item $\pi_B(\delta)$: the baseline policy parameterized by the distance to the geofence $\delta$.
	\item $\pi_{RTSA}(\alpha)$: the linear value function approximation based policy trained with RL with $R$ parameterized by $\alpha$.
\end{itemize}

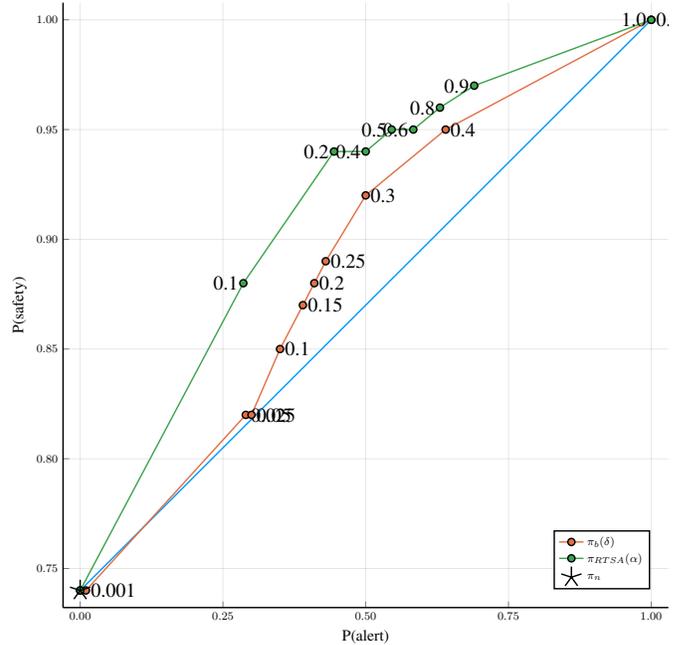
\begin{figure}[htbp]
\begin{center}
\begin{tikzpicture}[]
\begin{axis}[height = {165.1mm}, legend pos = {south east}, ylabel = {P(safety)}, xmin = {-0.03}, xmax = {1.03}, ymax = {1.0078}, xlabel = {P(alert)}, unbounded coords=jump,scaled x ticks = false,xlabel style = {font = {\fontsize{11 pt}{14.3 pt}\selectfont}, color = {rgb,1:red,0.00000000;green,0.00000000;blue,0.00000000}, draw opacity = 1.0, rotate = 0.0},xmajorgrids = true,xtick = {0.0,0.25,0.5,0.75,1.0},xticklabels = {$0.00$,$0.25$,$0.50$,$0.75$,$1.00$},xtick align = inside,xticklabel style = {font = {\fontsize{8 pt}{10.4 pt}\selectfont}, color = {rgb,1:red,0.00000000;green,0.00000000;blue,0.00000000}, draw opacity = 1.0, rotate = 0.0},x grid style = {color = {rgb,1:red,0.00000000;green,0.00000000;blue,0.00000000},
draw opacity = 0.1,
line width = 0.5,
solid},axis x line* = left,x axis line style = {color = {rgb,1:red,0.00000000;green,0.00000000;blue,0.00000000},
draw opacity = 1.0,
line width = 1,
solid},scaled y ticks = false,ylabel style = {font = {\fontsize{11 pt}{14.3 pt}\selectfont}, color = {rgb,1:red,0.00000000;green,0.00000000;blue,0.00000000}, draw opacity = 1.0, rotate = 0.0},ymajorgrids = true,ytick = {0.75,0.8,0.8500000000000001,0.9,0.9500000000000001,1.0},yticklabels = {$0.75$,$0.80$,$0.85$,$0.90$,$0.95$,$1.00$},ytick align = inside,yticklabel style = {font = {\fontsize{8 pt}{10.4 pt}\selectfont}, color = {rgb,1:red,0.00000000;green,0.00000000;blue,0.00000000}, draw opacity = 1.0, rotate = 0.0},y grid style = {color = {rgb,1:red,0.00000000;green,0.00000000;blue,0.00000000},
draw opacity = 0.1,
line width = 0.5,
solid},axis y line* = left,y axis line style = {color = {rgb,1:red,0.00000000;green,0.00000000;blue,0.00000000},
draw opacity = 1.0,
line width = 1,
solid},    xshift = 0.0mm,
    yshift = 0.0mm,
    axis background/.style={fill={rgb,1:red,1.00000000;green,1.00000000;blue,1.00000000}}
,legend style = {color = {rgb,1:red,0.00000000;green,0.00000000;blue,0.00000000},
draw opacity = 1.0,
line width = 1,
solid,fill = {rgb,1:red,1.00000000;green,1.00000000;blue,1.00000000},font = {\fontsize{8 pt}{10.4 pt}\selectfont}},colorbar style={title=}, ymin = {0.7322}, width = {165.1mm}]\addplot+ [color = {rgb,1:red,0.00000000;green,0.60560316;blue,0.97868012},
draw opacity = 1.0,
line width = 1,
solid,mark = none,
mark size = 2.0,
mark options = {
    color = {rgb,1:red,0.00000000;green,0.00000000;blue,0.00000000}, draw opacity = 1.0,
    fill = {rgb,1:red,0.00000000;green,0.60560316;blue,0.97868012}, fill opacity = 1.0,
    line width = 1,
    rotate = 0,
    solid
},forget plot]coordinates {
(0.0, 0.74)
(1.0, 1.0)
};
\addplot+ [color = {rgb,1:red,0.88887350;green,0.43564919;blue,0.27812294},
draw opacity = 1.0,
line width = 1,
solid,mark = *,
mark size = 2.5,
mark options = {
    color = {rgb,1:red,0.00000000;green,0.00000000;blue,0.00000000}, draw opacity = 1.0,
    fill = {rgb,1:red,0.88887350;green,0.43564919;blue,0.27812294}, fill opacity = 1.0,
    line width = 1,
    rotate = 0,
    solid
}]coordinates {
(0.0, 0.74)
(0.01, 0.74)
(0.29, 0.8200000000000001)
(0.3, 0.8200000000000001)
(0.35, 0.85)
(0.39, 0.87)
(0.41, 0.88)
(0.43, 0.89)
(0.5, 0.92)
(0.64, 0.9500000000000001)
(1.0, 1.0)
};
\addlegendentry{$\pi_b(\delta)$}
\node at (axis cs:0.0, 0.74) [right,
color={rgb,1:red,0.00000000;green,0.00000000;blue,0.00000000}, draw opacity=1.0,
rotate=0.0,
font={\fontsize{14 pt}{18.2 pt}\selectfont}
] {};
\node at (axis cs:0.01, 0.74) [right,
color={rgb,1:red,0.00000000;green,0.00000000;blue,0.00000000}, draw opacity=1.0,
rotate=0.0,
font={\fontsize{14 pt}{18.2 pt}\selectfont}
] {0.001};
\node at (axis cs:0.29, 0.8200000000000001) [right,
color={rgb,1:red,0.00000000;green,0.00000000;blue,0.00000000}, draw opacity=1.0,
rotate=0.0,
font={\fontsize{14 pt}{18.2 pt}\selectfont}
] {0.025};
\node at (axis cs:0.3, 0.8200000000000001) [right,
color={rgb,1:red,0.00000000;green,0.00000000;blue,0.00000000}, draw opacity=1.0,
rotate=0.0,
font={\fontsize{14 pt}{18.2 pt}\selectfont}
] {0.05};
\node at (axis cs:0.35, 0.85) [right,
color={rgb,1:red,0.00000000;green,0.00000000;blue,0.00000000}, draw opacity=1.0,
rotate=0.0,
font={\fontsize{14 pt}{18.2 pt}\selectfont}
] {0.1};
\node at (axis cs:0.39, 0.87) [right,
color={rgb,1:red,0.00000000;green,0.00000000;blue,0.00000000}, draw opacity=1.0,
rotate=0.0,
font={\fontsize{14 pt}{18.2 pt}\selectfont}
] {0.15};
\node at (axis cs:0.41, 0.88) [right,
color={rgb,1:red,0.00000000;green,0.00000000;blue,0.00000000}, draw opacity=1.0,
rotate=0.0,
font={\fontsize{14 pt}{18.2 pt}\selectfont}
] {0.2};
\node at (axis cs:0.43, 0.89) [right,
color={rgb,1:red,0.00000000;green,0.00000000;blue,0.00000000}, draw opacity=1.0,
rotate=0.0,
font={\fontsize{14 pt}{18.2 pt}\selectfont}
] {0.25};
\node at (axis cs:0.5, 0.92) [right,
color={rgb,1:red,0.00000000;green,0.00000000;blue,0.00000000}, draw opacity=1.0,
rotate=0.0,
font={\fontsize{14 pt}{18.2 pt}\selectfont}
] {0.3};
\node at (axis cs:0.64, 0.9500000000000001) [right,
color={rgb,1:red,0.00000000;green,0.00000000;blue,0.00000000}, draw opacity=1.0,
rotate=0.0,
font={\fontsize{14 pt}{18.2 pt}\selectfont}
] {0.4};
\node at (axis cs:1.0, 1.0) [right,
color={rgb,1:red,0.00000000;green,0.00000000;blue,0.00000000}, draw opacity=1.0,
rotate=0.0,
font={\fontsize{14 pt}{18.2 pt}\selectfont}
] {0.5};
\addplot+ [color = {rgb,1:red,0.24222430;green,0.64327509;blue,0.30444865},
draw opacity = 1.0,
line width = 1,
solid,mark = *,
mark size = 2.5,
mark options = {
    color = {rgb,1:red,0.00000000;green,0.00000000;blue,0.00000000}, draw opacity = 1.0,
    fill = {rgb,1:red,0.24222430;green,0.64327509;blue,0.30444865}, fill opacity = 1.0,
    line width = 1,
    rotate = 0,
    solid
}]coordinates {
(0.0, 0.74)
(0.2857142857142857, 0.88)
(0.4444444444444444, 0.94)
(0.5, 0.94)
(0.5454545454545454, 0.95)
(0.5833333333333334, 0.95)
(0.63, 0.96)
(0.69, 0.97)
(1.0, 1.0)
};
\addlegendentry{$\pi_{RTSA}(\alpha)$}
\node at (axis cs:0.0, 0.74) [left,
color={rgb,1:red,0.00000000;green,0.00000000;blue,0.00000000}, draw opacity=1.0,
rotate=0.0,
font={\fontsize{14 pt}{18.2 pt}\selectfont}
] {};
\node at (axis cs:0.2857142857142857, 0.88) [left,
color={rgb,1:red,0.00000000;green,0.00000000;blue,0.00000000}, draw opacity=1.0,
rotate=0.0,
font={\fontsize{14 pt}{18.2 pt}\selectfont}
] {0.1};
\node at (axis cs:0.4444444444444444, 0.94) [left,
color={rgb,1:red,0.00000000;green,0.00000000;blue,0.00000000}, draw opacity=1.0,
rotate=0.0,
font={\fontsize{14 pt}{18.2 pt}\selectfont}
] {0.2};
\node at (axis cs:0.5, 0.94) [left,
color={rgb,1:red,0.00000000;green,0.00000000;blue,0.00000000}, draw opacity=1.0,
rotate=0.0,
font={\fontsize{14 pt}{18.2 pt}\selectfont}
] {0.4};
\node at (axis cs:0.5454545454545454, 0.95) [left,
color={rgb,1:red,0.00000000;green,0.00000000;blue,0.00000000}, draw opacity=1.0,
rotate=0.0,
font={\fontsize{14 pt}{18.2 pt}\selectfont}
] {0.5};
\node at (axis cs:0.5833333333333334, 0.95) [left,
color={rgb,1:red,0.00000000;green,0.00000000;blue,0.00000000}, draw opacity=1.0,
rotate=0.0,
font={\fontsize{14 pt}{18.2 pt}\selectfont}
] {0.6};
\node at (axis cs:0.63, 0.96) [left,
color={rgb,1:red,0.00000000;green,0.00000000;blue,0.00000000}, draw opacity=1.0,
rotate=0.0,
font={\fontsize{14 pt}{18.2 pt}\selectfont}
] {0.8};
\node at (axis cs:0.69, 0.97) [left,
color={rgb,1:red,0.00000000;green,0.00000000;blue,0.00000000}, draw opacity=1.0,
rotate=0.0,
font={\fontsize{14 pt}{18.2 pt}\selectfont}
] {0.9};
\node at (axis cs:1.0, 1.0) [left,
color={rgb,1:red,0.00000000;green,0.00000000;blue,0.00000000}, draw opacity=1.0,
rotate=0.0,
font={\fontsize{14 pt}{18.2 pt}\selectfont}
] {1.0};
\addplot+[draw=none, color = {rgb,1:red,1.00000000;green,1.00000000;blue,1.00000000},
draw opacity = 1.0,
line width = 0,
solid,mark = star,
mark size = 7.5,
mark options = {
    color = {rgb,1:red,0.00000000;green,0.00000000;blue,0.00000000}, draw opacity = 1.0,
    fill = {rgb,1:red,0.76444018;green,0.44411178;blue,0.82429754}, fill opacity = 1.0,
    line width = 1,
    rotate = 0,
    solid
}] coordinates {
(0.0, 0.74)
};
\addlegendentry{$\pi_n$}
\end{axis}

\end{tikzpicture}
\end{center}
\caption{SOC Curves}
\label{fig:soc-curves}
\end{figure}

To compare the performance of the different policies we aggregated the outcome of each episode as confusion matrix consisting of the four possible outcomes determined by exiting $E$ or not while having deployed $\pi_r$ or not. 
For $\pi_B(\delta)$ and $\pi_{RTSA}(\alpha)$ we ran the episodes for different values of $\delta$ and $\alpha$ accordingly. 
For $\pi_n$, we obtained one single point which corresponds to letting the system run without RTSA.
\iffalse
\begin{table}[h]
\begin{center}
\begin{tabular}{l|l|l|}
\cline{2-3}
 & Safe & Unsafe \\ \hline
\multicolumn{1}{|l|}{Not Deployed} & a & b \\ \hline
\multicolumn{1}{|l|}{Deployed} & c & d \\ \hline
\end{tabular}
\end{center}
\end{table}
\chris{I dont like the format of this table, will try other things}

\fi
We then synthesized this information in Figure \ref{fig:soc-curves}. Where the $x$ axis represents the rate of alert and the $y$ axis the rate of conflict. 
Each point of the System Operating Characteristic curve corresponds to the probability of deploying the recovery controller (in the $x$ axis) and the probability of remaining within the safety envelope (in the $y$ axis).
%\begin{align}
%	\left ( \frac{c+d}{N}, 1-\frac{a+c}{N} \right )
%\end{align}

The closer the points in the curve of the associated system to the upper left corner, the better the system performs. We ideally want a system that deploys $\pi_r$ every time that it is needed to prevent exiting $E$ and that also prevents the aircraft from exiting $E$ every time the recovery controller is deployed.

%\begin{figure}[htbp]
%\centerline{\includegraphics[width=.45\textwidth]{img/SOC-placeholder}}
%\caption{SOC Curves}
%\label{fig:soc-curves}
%\end{figure}

As we can observe in this figure, the RL based policy consistently outperforms the baseline and also dominates the nominal controller. This result suggests that linear value function approximation based policies crafted with appropriate features can lead to a consistently better RTSA system than the natural human engineered approach.

\iffalse

\begin{table}[htbp]
\caption{Table Type Styles}
\begin{center}
\begin{tabular}{|c|c|c|c|}
\hline
\textbf{Table}&\multicolumn{3}{|c|}{\textbf{Table Column Head}} \\
\cline{2-4} 
\textbf{Head} & \textbf{\textit{Table column subhead}}& \textbf{\textit{Subhead}}& \textbf{\textit{Subhead}} \\
\hline
copy& More table copy$^{\mathrm{a}}$& &  \\
\hline
\multicolumn{4}{l}{$^{\mathrm{a}}$Sample of a Table footnote.}
\end{tabular}
\label{tab1}
\end{center}
\end{table}
\fi

\section{Conclusions}
\label{sec:conclusions}
Runtime safety assurance systems can enable the deployment of unverified components by monitoring them during operation to ensure the system remains within a predefined safety envelope.
RTSA systems should switch to a safe recovery controller when the vehicle is likely to exit the safety envelop. In this work we proposed a way to design the switching policy for RTSA and conducted experiments using a high fidelity flight simulator. Our experiments demonstrated the viability of this approach as it consistently exhibited superior performance to the human engineered baseline policy.

In this work we restricted our attention to terminal recovery controllers. Future work could investigate how to incorporate recovery controllers that can steer the vehicle to safe operational conditions and then return control to the nominal controller. 
Another direction for future work would be to incorporate the behavior of the system over time and analyze its safety as a closed-loop system with non-linear dynamics \cite{sidrane2019overt}.

\printbibliography

@String { dasc        = {Digital Avionics Systems Conference (DASC)} }

@String { gnc         = {AIAA Guidance, Navigation, and Control Conference (GNC)} }

@String { iclr        = {International Conference on Learning Representations} }

@String { ieeetits     = {IEEE Transactions on Intelligent Transportation Systems} }

@String { itsc        = {IEEE International Conference on Intelligent Transportation Systems (ITSC)} }

@String { jgcd        = {AIAA Journal of Guidance, Control, and Dynamics} }

@String { mit         = {Massachusetts Institute of Technology} }

@inproceedings{julian2016policy,
  title={Policy compression for aircraft collision avoidance systems},
  author={Julian, Kyle D and Lopez, Jessica and Brush, Jeffrey S and Owen, Michael P and Kochenderfer, Mykel J},
  booktitle=dasc,
  pages={1--10},
  year={2016}
}

@article{liu2019algorithms,
  title={Algorithms for Verifying Deep Neural Networks},
  author={Liu, Changliu and Arnon, Tomer and Lazarus, Christopher and Barrett, Clark and Kochenderfer, Mykel J},
  journal={arXiv preprint arXiv:1903.06758},
  year={2019}
}

@InProceedings{sidrane2019overt,
author = {Chelsea Sidrane and Mykel J. Kochenderfer},
title = {{OVERT}: {V}erification of nonlinear dynamical systems with neural network controllers via overapproximation},
booktitle = {Workshop on Safe Machine Learning, } # iclr,
year = {2019},
}

@article{setodynamic,
title = "Dynamic Control System Upgrade Using the Simplex Architecture",
author = "D. Seto and Krogh, {B. H.} and Sha, {Lui Raymond} and A. Chutinan",
year = "1998",
month = "8",
doi = "10.1109/37.710880",
language = "English (US)",
volume = "18",
pages = "72--80",
journal = "IEEE Control Systems",
number = "4",
}

@article{shasimplicity,
 	author = {Sha, Lui}, title = {Using Simplicity to Control Complexity}, year = {2001}, issue_date = {July 2001}, publisher = {IEEE Computer Society Press}, address = {Washington, DC, USA}, volume = {18}, number = {4}, doi = {10.1109/MS.2001.936213}, journal = {IEEE Software}, month = jul, pages = {20–28}, numpages = {9} }

@Book{dmu,
  Title                    = {Decision Making Under Uncertainty: Theory and Application},
  Author                   = {Mykel J. Kochenderfer},
  Publisher                = {MIT Press},
  Year                     = {2015}
}

@article{verifsurvey,
  author    = {Changliu Liu and
               Tomer Arnon and
               Christopher Lazarus and
               Clark Barrett and
               Mykel J. Kochenderfer},
  title     = {Algorithms for Verifying Deep Neural Networks},
  journal   = {CoRR},
  volume    = {abs/1903.06758},
  year      = {2019},
  archivePrefix = {arXiv},
  eprint    = {1903.06758},
  bibsource = {dblp computer science bibliography, https://dblp.org}
}

@Book{Kochenderfer2015,
author = {Mykel J. Kochenderfer},
publisher = {MIT Press},
title = {Decision Making Under Uncertainty: Theory and Application},
year = {2015},
chapter = {10}
}

@INPROCEEDINGS{Watkins92q-learning,
    author = {Christopher J. C. H. Watkins and Peter Dayan},
    title = {Q-learning},
    booktitle = {Machine Learning},
    year = {1992},
    pages = {279--292}
}

@article{dqn,
  author    = {Volodymyr Mnih and
               Koray Kavukcuoglu and
               David Silver and
               Alex Graves and
               Ioannis Antonoglou and
               Daan Wierstra and
               Martin A. Riedmiller},
  title     = {Playing Atari with Deep Reinforcement Learning},
  journal   = {CoRR},
  volume    = {abs/1312.5602},
  year      = {2013},
  archivePrefix = {arXiv},
  eprint    = {1312.5602},
  bibsource = {dblp computer science bibliography, https://dblp.org}
}

@InProceedings{Temizer2010,
author = {Selim Temizer and Mykel J. Kochenderfer and Leslie P. Kaelbling and Tomas Lozano-Perez and James K. Kuchar},
booktitle = gnc,
title = {Collision avoidance for unmanned aircraft using {M}arkov decision processes},
year = {2010},
address = {Toronto, Canada},
doi = {10.2514/6.2010-8040},
}

@InProceedings{Bouton2020,
author = {Maxime Bouton and Alireza Nakhaei and David Isele and Kikuo Fujimura and Mykel J. Kochenderfer},
booktitle = itsc,
title = {Reinforcement learning with iterative reasoning for merging in dense Traffic},
year = {2020},
}

@article{gliepaper, 
author = {Singh, Satinder and Jaakkola, Tommi and Littman, Michael L. and Szepesv\'{a}ri, Csaba}, title = {Convergence Results for Single-Step On-Policy Reinforcement-Learning Algorithms}, year = {2000}, issue_date = {March 2000}, publisher = {Kluwer Academic Publishers}, address = {USA}, volume = {38}, number = {3}, issn = {0885-6125}, doi = {10.1023/A:1007678930559}, journal = {Mach. Learn.}, month = mar, pages = {287–308}, numpages = {22} }

@Article{julianfire,
author = {Kyle D. Julian and Mykel J. Kochenderfer},
journal = jgcd,
title = {Distributed wildfire surveillance with autonomous aircraft using deep reinforcement learning},
year = {2019},
number = {8},
pages = {1768--1778},
volume = {42},
doi = {10.2514/1.G004106},
}

@Article{Menda2019,
author = {Kunal Menda and Yi-Chun Chen and Justin Grana and James W. Bono and Brendan D. Tracey and Mykel J. Kochenderfer and David H. Wolpert},
journal = ieeetits,
title = {Deep reinforcement learning for event-driven multi-agent decision processes},
year = {2019},
number = {4},
pages = {1259--1268},
volume = {20},
doi = {10.1109/TITS.2018.2848264},
}

@inproceedings{ge-suas,
author = {Ren, Liling and Castillo-Effen, Mauricio and Yu, Han and Yoon, Yongeun and Nakamura, Takuma and Johnson, Eric and Ippolito, Corey},
year = {2017},
month = {06},
pages = {},
title = {Small Unmanned Aircraft System Trajectory Modeling in Support of {UAS} Traffic Management},
doi = {10.2514/6.2017-4268}
}

@inbook{ge-simulator,
author = {Liling Ren and Mauricio Castillo-Effen and Han Yu and Yongeun Yoon and Takuma Nakamura and Eric N. Johnson and Corey A. Ippolito},
title = {Small Unmanned Aircraft System Trajectory Modeling in Support of {UAS} Traffic Management},
booktitle = {AIAA Aviation Technology, Integration, and Operations Conference},
chapter = {},
pages = {},
doi = {10.2514/6.2017-4268},
}

@article{alphago,
title	= {Mastering the game of Go with deep neural networks and tree search},
author	= {David Silver and Aja Huang and Christopher J. Maddison and Arthur Guez and Laurent Sifre and George van den Driessche and Julian Schrittwieser and Ioannis Antonoglou and Veda Panneershelvam and Marc Lanctot and Sander Dieleman and Dominik Grewe and John Nham and Nal Kalchbrenner and Ilya Sutskever and Timothy Lillicrap and Madeleine Leach and Koray Kavukcuoglu and Thore Graepel and Demis Hassabis},
year	= {2016},
journal	= {Nature},
pages	= {484--503},
volume	= {529}
}
\end{document}